%% file: sampling.tex
\documentclass[10pt,journal,compsoc]{IEEEtran}
\usepackage{wrapfig}

\usepackage{graphicx}
\usepackage{amssymb}
\usepackage{subcaption}
\usepackage{multicol}
\usepackage[cmex10]{amsmath}
\usepackage{amsfonts,mathtools}
\usepackage{url}
\usepackage{array}
\usepackage{url}
\usepackage{bm}
\usepackage{graphicx}
\usepackage{mathrsfs}
\usepackage{amsthm} 
\usepackage{bm}
\usepackage{tikz}
\usepackage{cite}
\usepackage{color,soul}
\usetikzlibrary{bayesnet}

\usepackage{adjustbox}
\usepackage{amssymb}
\usepackage{algorithm}
\usepackage{algorithmic}

\usepackage{authblk}
    \title{Dense Air Quality Maps Using Regressive Facility Location Based Drive By Sensing}
\author[1]{Charul Paliwal }
\author[1]{Pravesh Biyani}
\affil[1]{Indraprastha Institute of Information Technology, Delhi}

\input{preamble}
\input{preamble1}
 
\begin{document}
\maketitle
\input{abstract.tex}
\input{introduction.tex}
\input{related.tex}
\input{model.tex}

\input{result.tex}
\section{Acknowledgement}
The authors would like to thank Dr. Rishabh Iyer (University of Texas, Dallas) for his suggestions.
    \bibliographystyle{IEEEtran}
\bibliography{aaai22.bib}
\onecolumn
\input{supplementary}

\end{document}

%% file: preamble.tex
\def \X {\boldsymbol{\Theta}}

\def \x {\boldsymbol{x}}
\def \samp{\theta}

\def \Y {\boldsymbol{\mathcal{Y}}}

\def \Z {\mathbf{Z}}
\def \z {\mathbf{z}}
\def \n {\mathbf{n}}

\def \V {\mathbf{V}}
\def \U {\mathbf{U}}
\def \a {\mathbf{a}}
\def \b {\mathbf{b}}
\def \A {\mathbf{A}}
\def \G {\mathbf{G}}
\def \H {\mathbf{H}}
\def \C {\mathbf{C}}
\def \Bi {\mathbf{B}}
\def \S {\mathbf{S}}

\def \B {\mathcal{B}}
\def \T {\mathcal{T}}
\def \L {\mathcal{L}}
\def \M {\mathcal{M}}
\providecommand{\norm}[1]{\left \| #1 \right \|}

%% file: preamble1.tex
\def \X {\boldsymbol{\Theta}}

\def \x {\boldsymbol{x}}
\def \samp{\theta}

\def \Y {\boldsymbol{\mathcal{Y}}}

\def \Z {\mathbf{Z}}
\def \z {\mathbf{z}}
\def \n {\mathbf{n}}

\def \V {\mathbf{V}}
\def \U {\mathbf{U}}
\def \a {\mathbf{a}}
\def \b {\mathbf{b}}
\def \A {\mathbf{A}}
\def \G {\mathbf{G}}
\def \H {\mathbf{H}}
\def \C {\mathcal{C}}
\def \Bi {\mathbf{B}}
\def \Ci {\mathbf{C}}
\def \S {\mathbf{S}}
\def \D {\mathcal{D}}
\def \B {\mathcal{B}}
\def \T {\mathcal{T}}
\def \L {\mathcal{L}}
\def \M {\mathcal{M}}
\providecommand{\norm}[1]{\left \| #1 \right \|}
\def \Xib {\boldsymbol{\Xi}}
\def \mub {\boldsymbol{\mu}}

\def \I {\mathbf{I}}

%% file: abstract.tex
\begin{abstract}
Currently, fixed static sensing is a primary way to monitor environmental data like air quality in cities. However, to obtain a dense spatial coverage, a large number of static monitors are required, thereby making it a costly option. Dense spatiotemporal coverage can be achieved using only a fraction of static sensors by deploying them on the moving vehicles, known as the drive by sensing paradigm. The redundancy present in the air quality data can be exploited by processing the sparsely sampled data to impute the remaining unobserved data points using the matrix completion techniques. However, the accuracy of imputation is dependent on the extent to which the moving sensors capture the inherent structure of the air quality matrix.  Therefore, the challenge is to pick those set of paths (using vehicles) that perform representative sampling in space and time.  Most works in the current literature for vehicle subset selection focus on maximizing the spatiotemporal coverage by maximizing the number of samples for different locations and time stamps which is not an effective representative sampling strategy and results in noisy imputations. We present regressive facility location based drive by sensing, an efficient vehicle selection framework that incorporates the smoothness in neighboring locations and autoregressive time correlation while selecting the optimal set of vehicles for effective spatiotemporal sampling.
We show that the proposed drive by sensing problem is submodular, thereby lending itself to a greedy algorithm but with performance guarantees. We evaluate our framework on selecting a subset from the fleet of public transport in Delhi, India. We illustrate that the method proposed in the paper samples the representative spatiotemporal data against the baseline methods, reducing the extrapolation error on the simulated air quality data. Our method, therefore, has the potential to provide cost effective dense air quality maps.

  \end{abstract}

%% file: introduction.tex
\section{Introduction}
Air pollution has drawn attention in recent years because of its profound effect on people's health \cite{who:2019}. It is crucial to assess the pollution level in a region by monitoring the air quality levels and recommend strategies to combat pollution. There are significant factors that affect the air quality of an area, such as transportation, electricity, fuel uses, industrial parameters such as power plant emissions, etc. Hence, air quality monitoring is required to make strategies for emission control as well as verifying strategies that control emissions. \par

The most widely used method to monitor air quality is by using static sensing, that is mounting sensors at fixed locations. The temporal resolution of static sensing is high, i.e., the air quality data is available for almost all the sampled timestamps. However, the spatial coverage in a region depends on the number of sensors installed which is generally limited owing to the cost constraints. We can leverage the spatial and temporal correlation in the air quality data to perform cost-effective spatiotemporal monitoring. We can share/multiplex an air quality sensor over multiple locations and perform sequential temporal sensing without losing air quality information. Moving sensor paradigm can be a candidate to obtain such dense AQ (Air Quality) map without needing an expensive setup of hundreds of static monitors \cite{anjomshoaa2018city,hasenfratz2015deriving}. With the technological progress in AQ sensing, it is now possible to put these monitors in buses and other public transit and perform spatiotemporal AQ sensing. However, the moving sensors reading would be spatially and temporally sparse and incomplete because of the unavailability of the moving vehicle across all the spatial locations and timestamps.

Thanks to the inherent temporal and spatial redundancy available in the air quality data, one can potentially impute the remaining (incomplete) data by utilising the matrix completion techniques \cite{asif2016matrix,ownwork} and obtain a dense air quality map in a fraction of the corresponding fixed sensing cost. However, the quality of imputation crucially relies on the  "quality” of drive by sensing and the choice of "routes” that vehicles take to sample the air quality. 



The problem of dense air quality maps using drive by sensing 
is defined as: Given a binary bus occupancy data  ${\Y} \in \{0,1\}^{|\L|\times |\T|\times |\B|}$, where $\L$ denotes the set of locations, $\T$ denotes the set of timestamps and  $\B$ denotes the buses. The entry $y_{l,t,b}=1$ if the bus b samples the location $l$ in the time stamp $t$. The aim is to pick the best subset of buses $\M \in \B$ that perform representative sampling for the spatiotemporal AQ matrix. Subsequently, by appropriately modeling the spatial and temporal structure present in the AQ data sampled by the set of buses $\M$, one can extrapolate/impute the missing data in the spatiotemporal matrix. In a nutshell, the aim is to pick those buses that exploit the spatiotemporal structure in the air quality data resulting in an effective imputation and thereby creating dense anytime-anywhere AQ map.

Limited work on selecting the optimal set of vehicles for sensing spatiotemporal data models each entry in the spatiotemporal matrix as independent and tries to maximize the number of entries in the matrix \cite{related1, related2,related4}. Considering all the locations as independent ignores the correlation across locations, thereby highlighting the sub-optimality of such an approach. Indeed, sampling two consecutive neighborhood locations are less effective than sampling the diverse spread across locations in an area.

Our work exploits the smoothness in the spatial locations to select buses that sample diverse and representative sets of locations. Further, the air quality data for a given location is smooth in time, i.e., the sensor data varies slowly in time with decreasing correlation as the interval increases. Thereby, sampling the consecutive timestamps of a site is a less effective strategy than sampling at distant time stamps. The proposed Regressive Facility Location (RFL) framework encapsulates the temporal smoothness using an autoregressive time series structure to sample the spatiotemporal data. Also, since the future temporal data is not available to create the dense map at a particular timestamp, we use the causal temporal smoothness in the proposed RFL framework. This ensures that if a location $l$ is sampled at a given time, other buses can bypass sampling location $l$ and the nearby locations in the subsequent neighboring future time stamps. The Regressive Facility Location (RFL) framework proposes to encapsulate the slowly varying temporal data pattern into the facility location framework and select the buses such that the sampled locations will be representative of the area and will be diverse across all time stamps. 

 Practically it is not possible to access the real-world air quality data without deploying the sensors on the selected set of buses. Therefore, to evaluate the performance of the proposed RFL drive by sensing framework, we simulate real-world air quality data. Further, we obtain the dense air quality maps using the matrix completion framework from the sampled spatio-temporal data and observe that RFL provide more accurate dense air quality maps. 
 The overall contribution of our work are:
\begin{enumerate}
\item We propose Regressive Facility Location (RFL), a novel method to select a fleet of vehicles (buses) for drive by sensing that incorporates the smoothness of the sensor reading across space as well as time to provide effective data sampling. 
\item We show that the proposed RFL algorithm gain is submodular, therefore a greedy heuristic provides a solution within an approximation ratio of  $(1-\frac{1}{e})$. 
\item We simulate the smooth data over space and time and sample the area using the proposed drive by sensing framework. To evaluate the performance of the proposed framework, we create dense maps using a matrix completion framework from the sampled data. 
\end{enumerate}

%% file: related.tex
\subsection{Notations}
A tensor is represented by $\Y$. Matrix is denoted by $\mathbf{Y}$. A set is denoted by $\mathcal{Y}$, the number of element in a set is denoted by $Y$ where $Y=|\mathcal{Y}|$, the element of the set is denoted by $y$. 

\section{Background Review}

\subsection{Vehicle Subset Selection}
There has been limited work on selecting the optimal set of vehicles for sensing spatiotemporal data. Authors in \cite{related4} propose mobile sensor placement for vehicles to maximize coverage. Authors in \cite{related1} propose a drive by sensing framework for taxis and buses. Authors in \cite{related2} proposed Point of interest oriented (POIs) Bus Selection algorithm to select buses where the coverage of a bus is defined in terms of the POIs. Authors in \cite{related3} proposed the optimal placement of reference monitors to make mobile sensors
k-hop calibrable. However, the mobile sensors are fixed and reference monitors are optimally selected. All these framework maps the problem of vehicle subset selection to the maximum cover problem or set cover problem to maximize the spatiotemporal coverage. 
\subsection{Greedy Submodular Maximization}
\textbf{Definition 1:} A function f is monotonically non decreasing  if $\forall \,\,\,\mathcal{C} \subseteq \mathcal{D} $
\[f(\mathcal{C}) \le f(\mathcal{D}) \]
\textbf{Definition 2:} A function f: $2^{\mathcal{B}} \rightarrow \mathbf{R} $ is submodular if $\forall \,\,\,\mathcal{C} \subseteq \mathcal{D} \subseteq \mathcal{B}$ and every  $b \in \mathcal{B} \backslash \mathcal{D}$
\[f(\mathcal{C} \cup b)-f(\mathcal{C}) \ge f(\mathcal{D} \cup b)-f(\mathcal{D})  \]
Given a submodular set function $f$ , maximization of $f$ over all subsets of size at most $k$ of the ground set $\mathcal{G}$, i.e. $|\mathcal{G}|=k$,

\begin{equation}
f(\mathcal{V})= \max\limits_{ \mathcal{V}: |\mathcal{V}|\leq k} f(\mathcal{V})
\end{equation}
is an NP-hard problem \cite{fujishige2005submodular}. A monotone non decreasing submodular function solution can be approximate by an greedy algorithm within  $(1-\frac{1}{e})$ of the global maximum \cite{sub1, sub2}.

\subsection{Generating Smooth Graph Signals}
 To generate a smooth graph signal across nodes, adjacency matrix can be used.
Consider a graph with $\L$ set of nodes/locations, the weighted adjacency matrix can be defined as $\G \in \mathbf{R}^{L \times L}$ can be decomposed as. 
\[\G= \U \Sigma \U^{T}\]
The graph Fourier transform of a signal $\x \in \mathbf{R}^{L}$ is given by $\hat{\x}$.
\[\hat{\x}= \U^{-1} \x\]
The actual signal can be expressed as
\[\x= \U \hat{\x}\]

The signal is smooth in the time domain if the signal is bandlimited in the frequency domain \cite{spectral_proxy,mostimp,check5,check6,distance_based} hence,
\[\hat{\x}_{k}=0 \,\,\, \forall k \geq m\]
Therefore
\begin{equation}
	\x= \U_{(m)} \hat{\x}
	\label{eq:related1}
\end{equation}

where $ \U_{(m)} \in \mathbf{R}^{L \times m}$ represents the first $m$ eigenvectors of the matrix $\G$. \par
The signal $\hat{\x}$ can be sampled from random normal distribution as described in the literature \cite{mostimp,distance_based,check6}. 
Then using the signal $\hat{\x}$ and $\U$, actual smooth signal can be constructed over the nodes of the graph via Eq. \ref{eq:related1}. The signal generated will be smooth on the locations only. We simulate real-world air quality data that is smooth in space and time.

\subsection{Dense maps using Missing data imputation}
Air quality data exhibits both spatial and temporal correlation thereby generating redundancy (not high rank nature) \cite{sampson2011pragmatic,min2011real}. Low rank matrix completion has been proposed to estimate the missing spatiotemporal data \cite{asif2016matrix,mi1}. Further state space model for incorporating the temporal evolution in the matrix completion framework is also proposed in the literature \cite{varst,compare1,ownwork}. 
However, dense air quality maps cannot be computed using the traditional matrix completion framework for the cold start locations. Cold start locations refers to the set of locations that are never sampled by the selected set of buses. Similarity matrix can be exploited to impute the data for the cold start locations \cite{cold1,cold2}. 

%% file: model.tex
\section{Bus Selection Framework}
Bus selection for drive by sensing is defined as follows: We are given a binary bus occupancy tensor ${\Y} \in \{0,1\}^{|\L|\times |\T|\times  |\B|}$, where $ \L$ is the set of locations, $\T$ is the set of timestamps, and  $\B$ is the entire set of buses. Buses run on a specific route with a time schedule. The entry $y_{l,t,b}=1$ denotes the availability of the sensor reading placed at bus $b$ for a particular location $l$ and time slot $t$.  Our work aims to pick $k$ buses from the entire set of buses $B$ for drive by sensing, where  $\M \subset \B$ represents the selected set of buses. 
For a subset of buses $\M \subset \B$, we define a binary sampling matrix $\X \in \{0,1\}^{L\times T}$ as 
\begin{equation}
\X(\M)= \max(\sum\limits_{ k \in \M} \Y_{l,t,k},1)\,\,\,\,\, \forall \,\,(l \in \L),  \,\,\, \forall \,\,(t \in \T)
\label{eq:samp_mtx}
\end{equation}
The distance between two locations $i$ and $j$ are defined as $\mathbf{D}_{i,j}$. We then define the normalized distance between two locations $\textbf{ND}_{i,j}=\frac{\mathbf{D}_{i,j}}{\max{(\mathbf{D})}}$. All the entries in $\textbf{ND}$ are between 0 and 1. The smaller the normalized distance between two locations, the higher is the similarity. To incorporate the smoothness in the spatial locations, we define the similarity between locations in $\S \in \mathbf{R}^{L \times L}$ as:
\begin{equation}
\S_{i,j} =1-\textbf{ND}_{i,j }
\label{eq:sim}
\end{equation}

\subsection{Baseline methods}
\subsubsection{Max Coverage (MC) }
\label{sec: Maxcover}
We model our problem as the max coverage problem where each entry in the sampling matrix $\X$ is considered as an element\cite{related4}. We have the set of buses, each bus $e$ samples some elements in the matrix $\X(e)$. Max Coverage framework selects the $k$ number of buses, set $\M \subset \B$ where $|\M| =k$ such that entries in the sampling matrix $\X (\M)$ is maximized. The maximum coverage problem is NP-hard and submodular; there exists a greedy heuristic that provides a solution within an approximation ratio of  $(1-\frac{1}{e})$\cite{hochbaum1996approximating}. We use Percentage Coverage as a measure for the Max Coverage Algorithm. The gain for Max Coverage to maximize the occupancy of the selected buses $\M$ is defined as
\begin{equation}
PC(\M)= \frac{  \sum\limits_{i \in \L, j \in \T} \X(\M)_{i,j}}{ L \times T}*100
\label{eq:PC}
\end{equation}

%
%
%
%
%
%
%
%

\subsubsection{Max Coverage over Location (MCL)}
\label{sec: MaxcoverLoc}
We model our problem as the max coverage over location problem where each location is considered as an element. We have the set of buses, each bus samples some locations over all the timestamps. Max Coverage over location framework selects the $k$ number of buses, set $\M \subset \B$ where $|M| =k$ such that maximum locations are sampled. In max coverage over location, we are maximizing the number of locations sampled by the selected buses, while in max Coverage framework, we are maximizing the entries of the sampling matrix $\X$ to increase the coverage over location as well as time. We use Percentage Stop coverage as a measure for the  Max Coverage over Location algorithm as shown in Eq. \ref{eq:PSC}. 
\begin{equation}
PSC(\M)= \frac{  \sum\limits_{i \in \L} max\{ \sum\limits_{ k \in \M, j \in \T} \Y_{i,j,k},1\}}{  L}*100
\label{eq:PSC}
\end{equation}
%
%
%
%
%
%
%

\subsection{Facility Location over space (FLS)}
\label{sec:FLS}

Max coverage and max coverage location treat every location as independent while selecting the set of buses, which sample the spatiotemporal matrix. However, a correlation exists across the locations readings; the sensor readings will be smooth over neighboring locations.  We thereby use Facility Location over space (FLS) to model the smoothness in the locations using a similarity matrix.\par

For the selected set of buses $\M$, the sampling matrix is defined in Eq.\ref{eq:samp_mtx}. For a time stamp $t$, the sampled locations by the set of buses $\M$ is defined as $\samp_t(\M)$, 
\[ \samp_t (\M)= \{i\} \,\,\,\forall \,\,\,\X_{i,t}(\M)|_{=1}\]
\begin{equation}
FLS (\M)= \frac{\sum\limits_{ t \in \T} \sum\limits_{l\in \L} \pi^{l}_{t} (\M )}{ L \times T} 
\label{eq:FLS}
\end{equation}
\begin{equation}
\pi^{l}_{t} (\M )=\max\limits_{\forall m \in \samp_t (\M)}(\S_{l,m} )
\label{eq:pifls}
\end{equation}

We compute the maximum pairwise similarity between all the locations ($i \in \L$) and the sampled set locations denoted by $\samp_t (\M)$ for a timestamp $t$. The gain defined in Eq. \ref{eq:FLS} is maximized when the pairwise similarities between the locations and the nearest chosen location in the selected set of buses are maximized for all the timestamps. This ensures that we pick the buses that sample the locations that are representative of the entire area for all the timestamps.


\textbf{Theorem 1}: The Function defined in Eq. \ref{eq:FLS} is monotone submodular. Therefore a greedy heuristic  provides a solution within an approximation ratio of  $(1-\frac{1}{e})$.\\
Proof:  The proof is carried out in the appendix.

%
%
%
%
%
%
%
%
%
\begin{figure}
	\centering
	\includegraphics[clip, trim=0.2cm 0.2cm 0.5cm 0.2cm,width=0.6\linewidth]{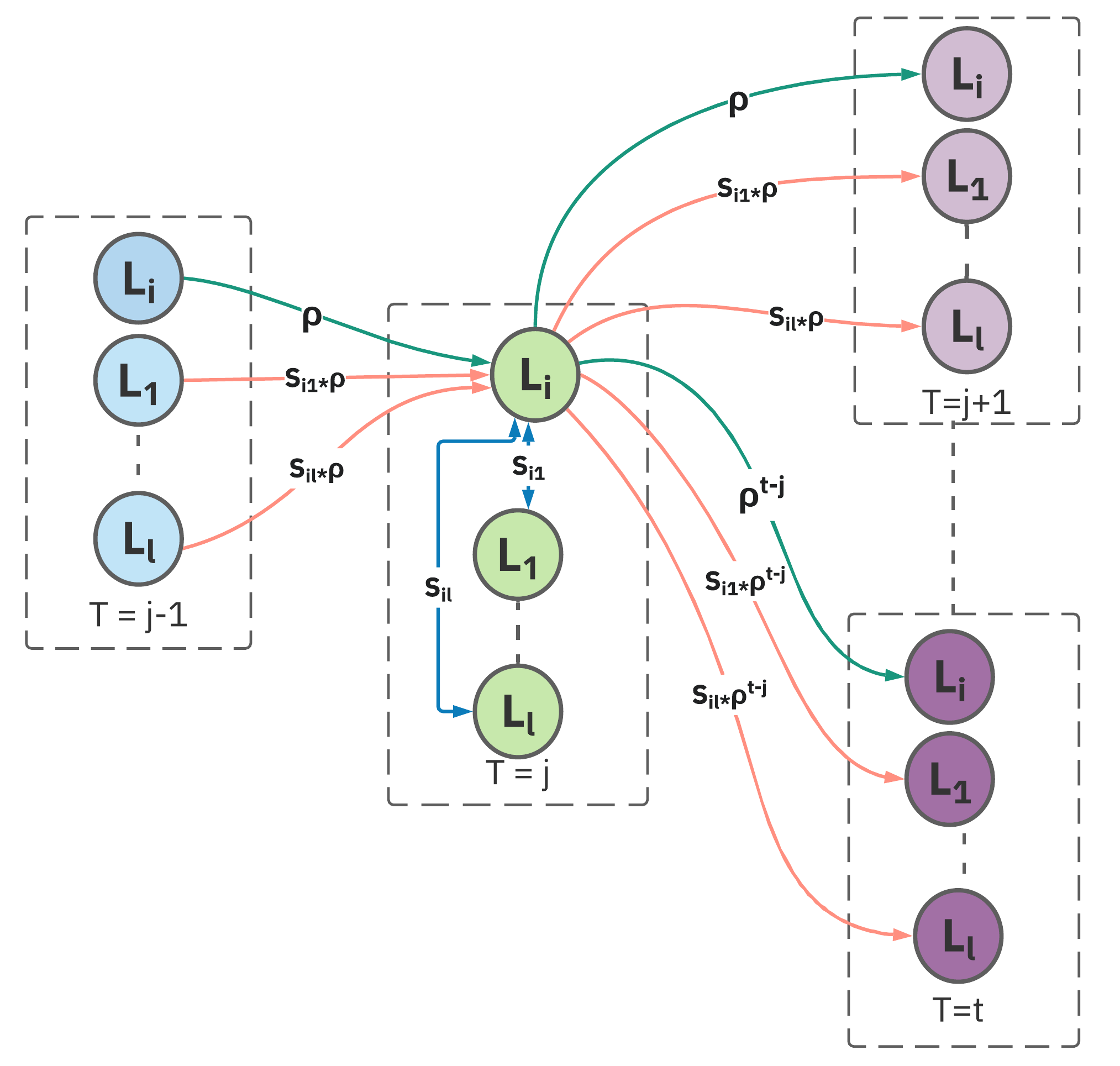}
	\caption{Spatial and temporal similarity}
	\label{fig:model}
\end{figure}
\begin{figure}
	\centering

	\includegraphics[clip, trim=2.5cm 0.2cm 2.5cm 0.2cm,width=1\linewidth]{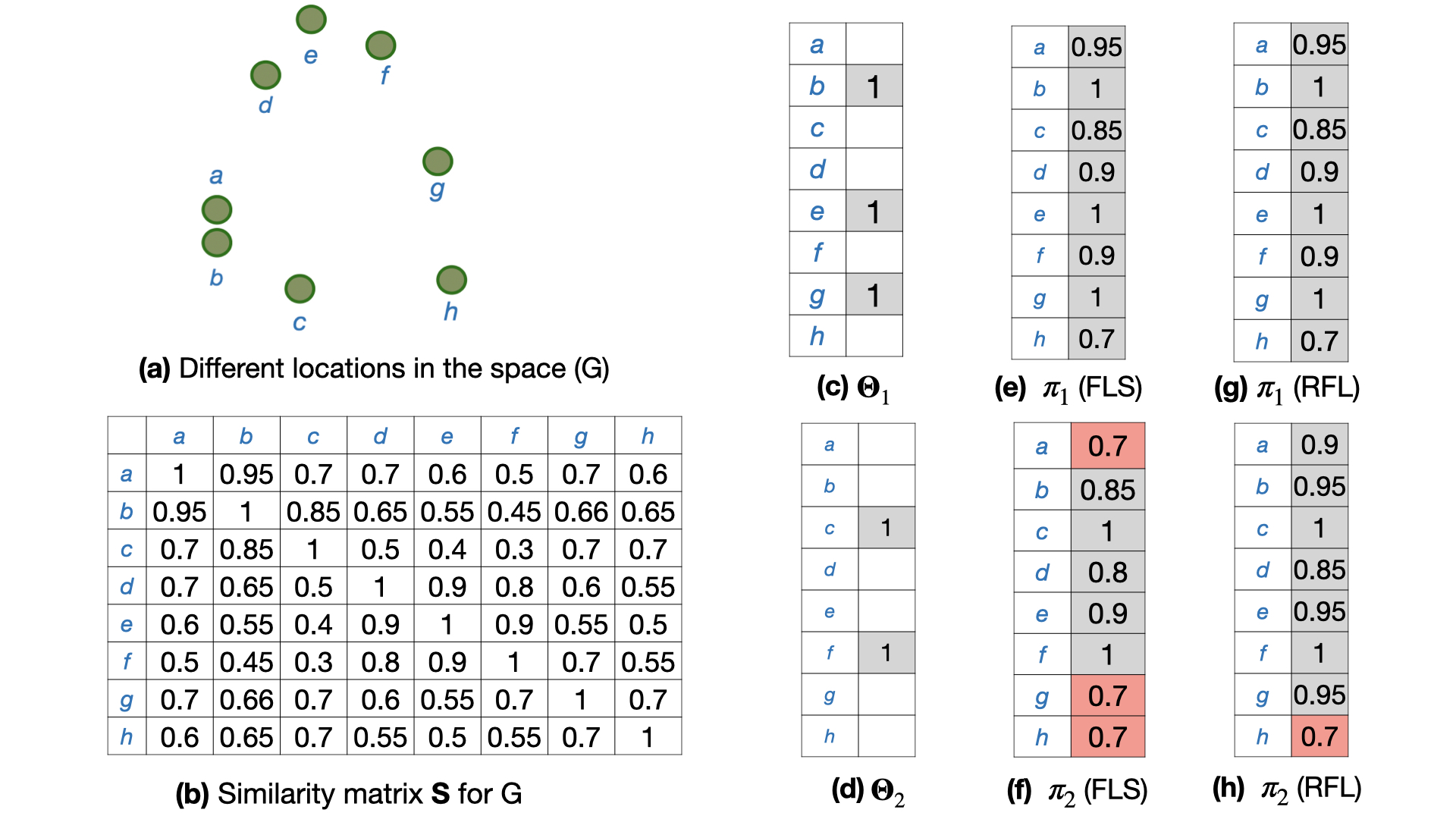}
	\caption{\small (a) Represents the locations of an example graph G, (b) The Similarity matrix $S$ for Graph G, (c)-(d) Sampling matrix $\X$ for time t=1 and 2, (e)-(f) $\pi_{t}$ for FLS, (g)-(h) $\pi_{t}$ for RFL. }
	\label{fig:exam_rfl}
\end{figure}

\subsection{Regressive Facility Location (RFL)}
FLS selects the buses that sample representative locations across all timestamps. However, fails to capture the temporal correlation while selecting the set of buses and treat all timestamps as independent. The sensor observations of a location over time are not independent and varies slowly over time. The FLS framework can be modified to incorporate the space and time similarity as shown in Fig. \ref{fig:model}.\par 
The modified framework incorporates the  temporal causal similarity  i.e., the future time data is not available to infer the previous time stamps using $\rho$. The Facility location incorporating the space and time causal similarity is called Facility Location over Space-Time (FLST). The similarity coefficients for location $i$ and timestamp $j$ is denoted in Fig. \ref{fig:model}. Blue edges in the graph denote the similarity between different locations for a particular timestamp $j$. Note that $\rho=0$ is a special case for FLS, and only blue edges are used for similarity in the FLS framework. 
FLST models the inter-temporal similarity for a particular location using $\rho$ and is denoted by green edges in Fig \ref{fig:model}. The inter location and inter-temporal similarity are denoted by orange edges. Directed edges represent the similarity from one node to the other. Note that since we are modeling the temporal causal similarity, reading at location $i$ is relevant only for the locations at future timestamps. 
The gain for the FLST is defined as 
\begin{equation}
FLST (\M)= \frac{\sum\limits_{ t \in \T} \sum\limits_{l \in \L} \pi_{t}^{l}(\M)}{ L \times T} 
\label{eq:FLST}
\end{equation}
\begin{equation}
\pi_{t}^{l}(\M)=\max\limits_{ j, m\in \X (\M)} (\S_{l,m} *\mathbf{T}_{t,j})
\label{eq:piflst}
\end{equation}
where the Temporal causal similarity is defined as 
\begin{equation}
\mathbf{T}_{t,j}= 
\begin{cases}
\rho^{t-j},&  t\geq j\\
0,              & t< j
\end{cases}
\end{equation}
It is computationally expensive to compute the Eq. \ref{eq:FLST} since the comparison for every location and time is done for all the sampled data $\X$. Therefore, we propose a fast algorithm that optimizes the gain defined in Eq. \ref{eq:FLST}, with a reduced computational complexity. 

Let $\samp_t$ be the set of locations that corresponds to sampled locations in the sampling matrix $\X$ for time stamp $t$. For $\pi_{0}=0$, $\rho \in [0,1]$, the regressive facility location gain is defined by 
\begin{equation}
RFL(\M )=  \frac{  \sum\limits_{ t \in \T} \sum\limits_{l \in \L}\pi_{t}^{l}(\M) }{L \times T}
\label{eq:RFL}
\end{equation}
\begin{equation}
\pi_{t}^{l}(\M)=\max((\S_{l,m})_{\forall m \in \samp_t(\M)} , \rho \pi_{t-1}^{l}(\M))  \,\,\,\forall l \in \L
\label{eq:pirfl}
\end{equation}
\textbf{Corollary 1}: FLST gain defined via Eq. \ref{eq:piflst} is equivalent to the RFL gain defined via Eq. \ref{eq:pirfl}. \\
Proof:  The proof is carried out in the appendix.\par
RFL encapsulates the temporal correlation to calculate the gain for $\pi_{t}^{l}$ by using $\pi_{t-1}^{l}$. The idea for incorporating the temporal sampling diversity is if the location is sampled at time $t_1$, then the subsequent neighboring time sampling would not be the best informative sampling as the data is smooth over time. In RFL, we incorporated the smoothness over time as shown in Fig. \ref{fig:exam_rfl}. Given that a location $g$ is sampled at time $t=1$, then the $\pi_{2}$ gain in the next timestamp is lower in FLS than RFL. 
Moreover, suppose we have two adjacent locations $a$ and $b$, that are neighboring nodes with high similarity. In that case, sampling the location $b$ at timestamp $t_1$ will also provide information for the location $a$ at subsequent neighboring time stamps as shown in Fig. \ref{fig:exam_rfl}. The greedy approach for the Facility location over space is shown in Algorithm \ref{Algo 2}.  \par 
\textbf{Theorem 2}: The Function defined in Eq. \ref{eq:RFL} is monotone submodular. Therefore a greedy heuristic  provides a solution within an approximation ratio of  $(1-\frac{1}{e})$.\\
Proof:  The proof is carried out in the appendix.\par
\textbf{Computational Complexity}: For each $l \in \L$ and $t \in \T$ computing $\pi_{t}^{l}(\M)$ via Eq. \ref{eq:pifls} requires $\Lambda$ comparisons where $\Lambda < L$ hence the computational complexity for each iteration of the greedy algorithm for all the buses in FLS is $O(\Lambda LTB)$. 
For each $l \in \L$ and $t \in \T$ computing $\pi_{t}^{l}(\M)$ via Eq. \ref{eq:piflst} requires $\Lambda T$ comparisons, hence a cost of  $O(\Lambda LT^2B)$ for FLST. 
We reduce the computation complexity of RFL by computing $\pi_{t}^{l}(\M)$ via Eq. \ref{eq:pirfl} to  $O(\Lambda LTB)$. 

\begin{algorithm}
	\caption{Regressive Facility Location}

	\label{alg:temp_fac_loc}
	\textbf{Input}: $k,\,\,\, \B, \,\,\, \Y, \,\, \rho,\,\,\, \S$\\
	\textbf{Output}: $\M \subset \B$ of size $k$
	
	\textbf{Initialization}:$\M\leftarrow  \phi$; $\pi_{0}=0$
	
	\begin{algorithmic}[1] 
		
		\FOR{$i$ =1 to $k$}
		\FOR{each $e \in \B$ }
		\STATE $\X(\M \cup e)= \max(\sum\limits_{ k \in \M \cup e} \Y_{i,j,k},1)$
		\FOR{each $t \in \T$}
		\STATE	$ \samp_t (\M \cup e)= \{i\} \,\,\,\forall \,\,\,\X_{i,t}(\M \cup e)|_{=1}$
		\STATE 	Compute $\pi_{t}^{l}(\M \cup e)$ using Eq. \ref{eq:pirfl}  $\,\,\,(\forall l \in \L)$

		\ENDFOR
		\STATE  $f(\M \cup e)=    \frac{1}{L \times T}\sum_{t=1}^{T} \sum_{l=1}^{L}\pi_{t}^{l}(\M \cup e)  $
		
		\ENDFOR
		\STATE $	e^{*}= \arg \max_{e} f(\M \cup e)$
		\STATE $	\M \leftarrow \M \cup \{e^{*}\}$

		\ENDFOR
	\end{algorithmic}
	\label{Algo 2}
\end{algorithm}

%% file: result.tex
\section{Experimentation}
\subsection{Simulating Real World AQ data}
\subsubsection{Similarity matrix}
\label{sec:res_sim}
To simulate the real world AQ data on the set of location $\L$ and timestamps $\T$, we learn the similarity matrix based on the static real world of Delhi \cite{cpcb}. It contains the AQ data for 33 locations for 3 months. 
We define the entry of the Similarity matrix $\G_{i,j}= \exp^{(-\lambda d_{i,j})}$ where $d_{i,j}$ is the distance between the two locations $i$ and $j$. We learn the parameter $\lambda$ using linear regression and it is observed to be 0.07676 for the AQ data \cite{cpcb}. We then compute the similarity matrix $\G $ for the set of location $\L$ based on the actual distance between locations $d_{i,j}$. We learn the temporal similarity matrix from the data \cite{cpcb} and use it as a temporal similarity matrix $\H$. 



\begin{figure*}[ht!]
    \centering
    \begin{subfigure}{.4\linewidth}
        \includegraphics[scale=0.45]{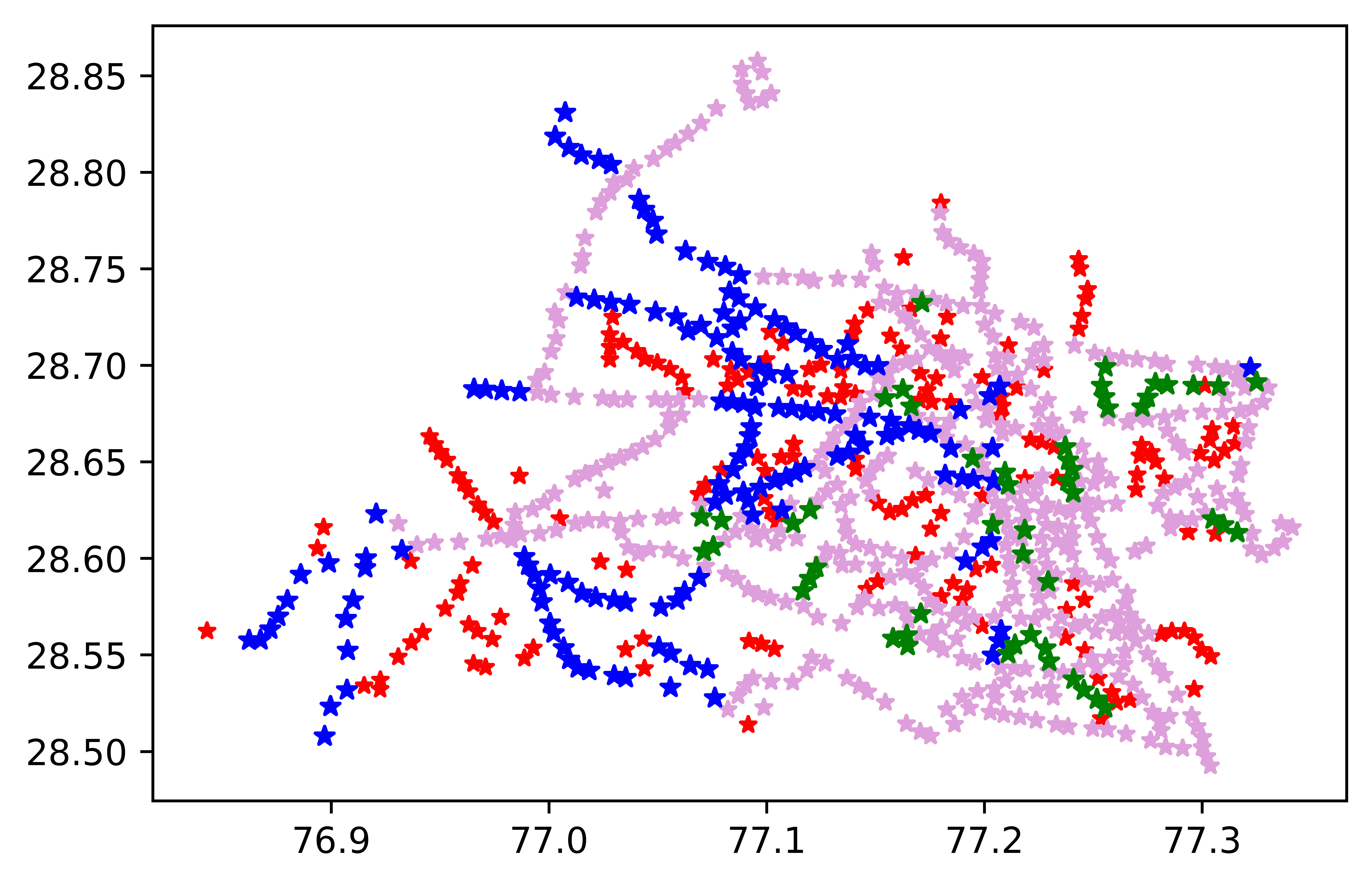}
        \caption{RFL vs MC for atleast 1 time stamp }
    \end{subfigure}
    \hskip1em
    \begin{subfigure}{.4\linewidth}
        \includegraphics[scale=0.45]{image/plot_comp_max_cover.png}
        \caption{RFL vs MC for atleast 10 time stamp}
    \end{subfigure}
    \begin{subfigure}{.4\linewidth}
        \includegraphics[scale=0.45]{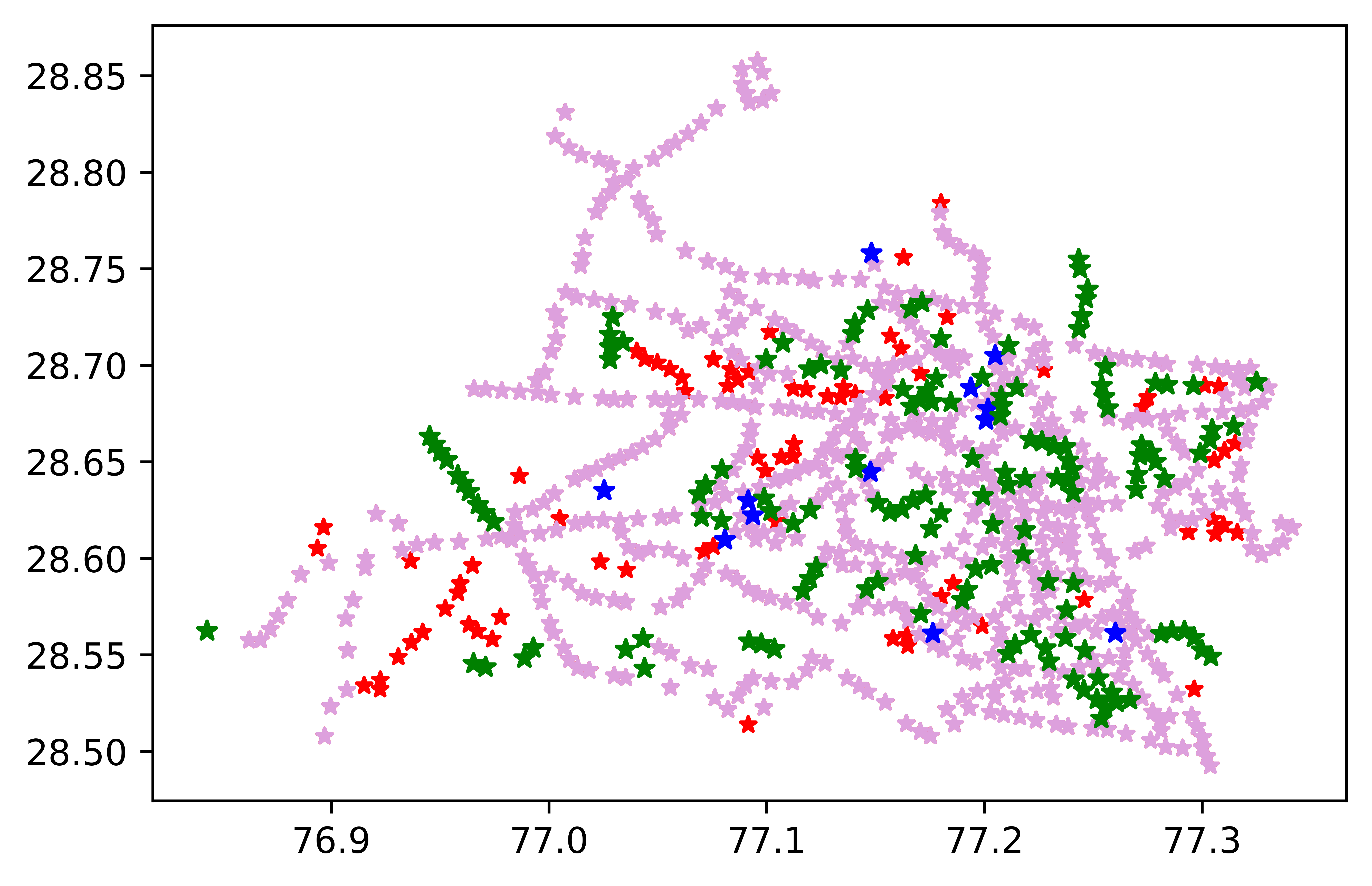}
        \caption{RFL vs MCL for atleast 1 time stamp }
    \end{subfigure}
    \hskip1em
    \begin{subfigure}{.4\linewidth}
        \includegraphics[scale=0.45]{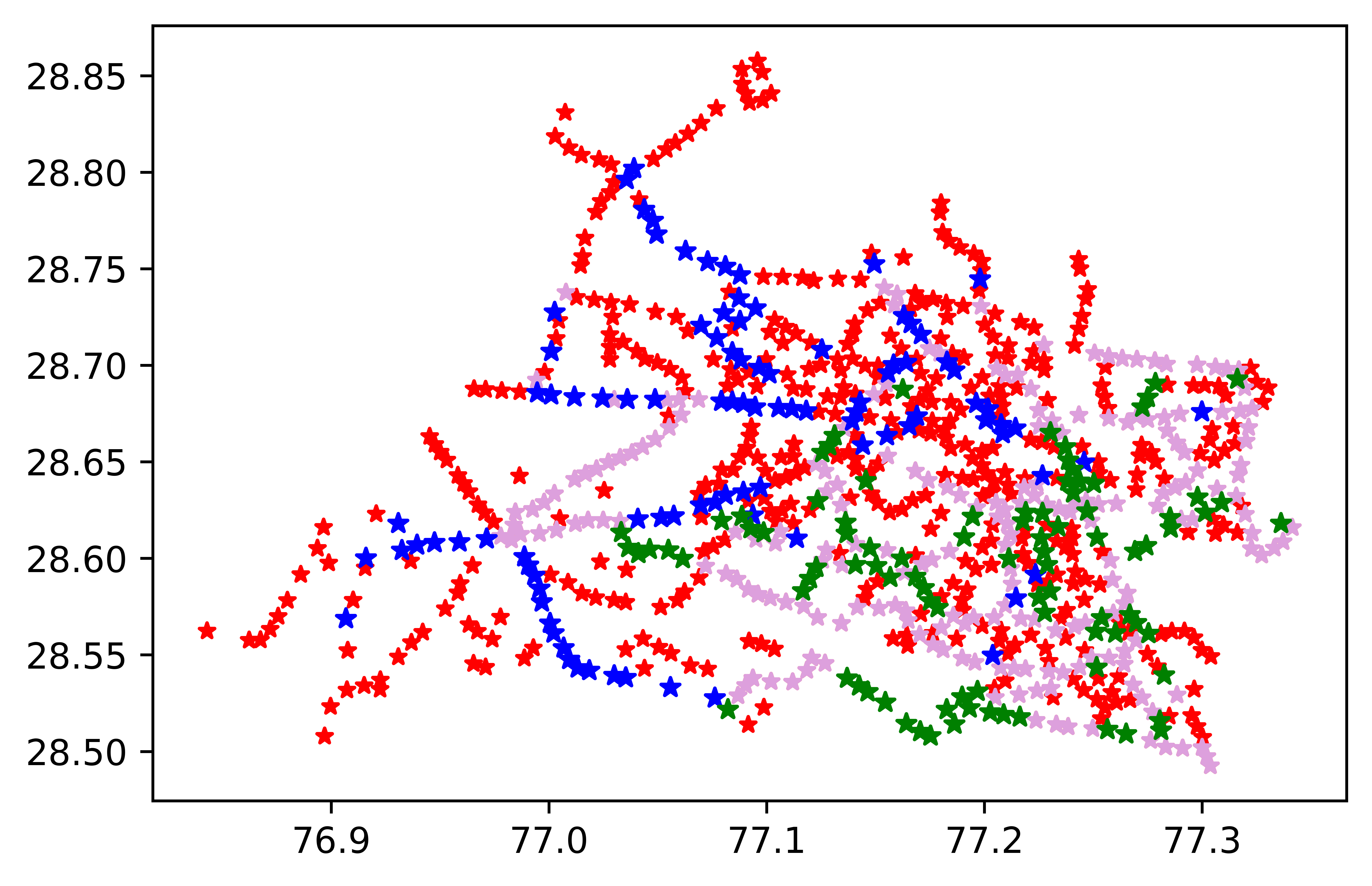}
        \caption{RFL vs MCL for atleast 10 time stamp}
    \end{subfigure}
    \begin{subfigure}{.4\linewidth}
        \includegraphics[scale=0.45]{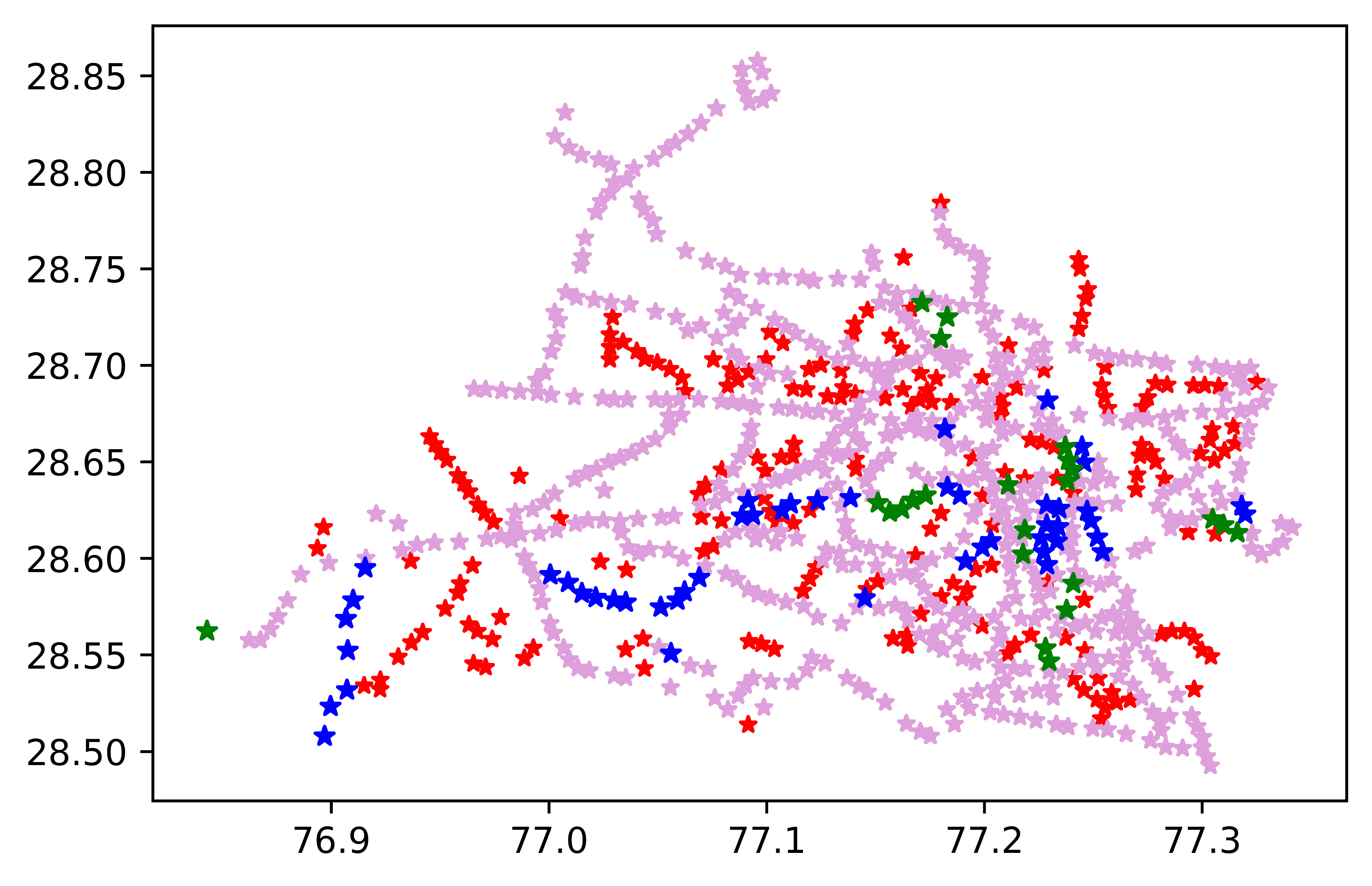}
        \caption{RFL vs FLS for atleast 1 time stamp }
    \end{subfigure}
    \hskip1em
    \begin{subfigure}{.4\linewidth}
        \includegraphics[scale=0.45]{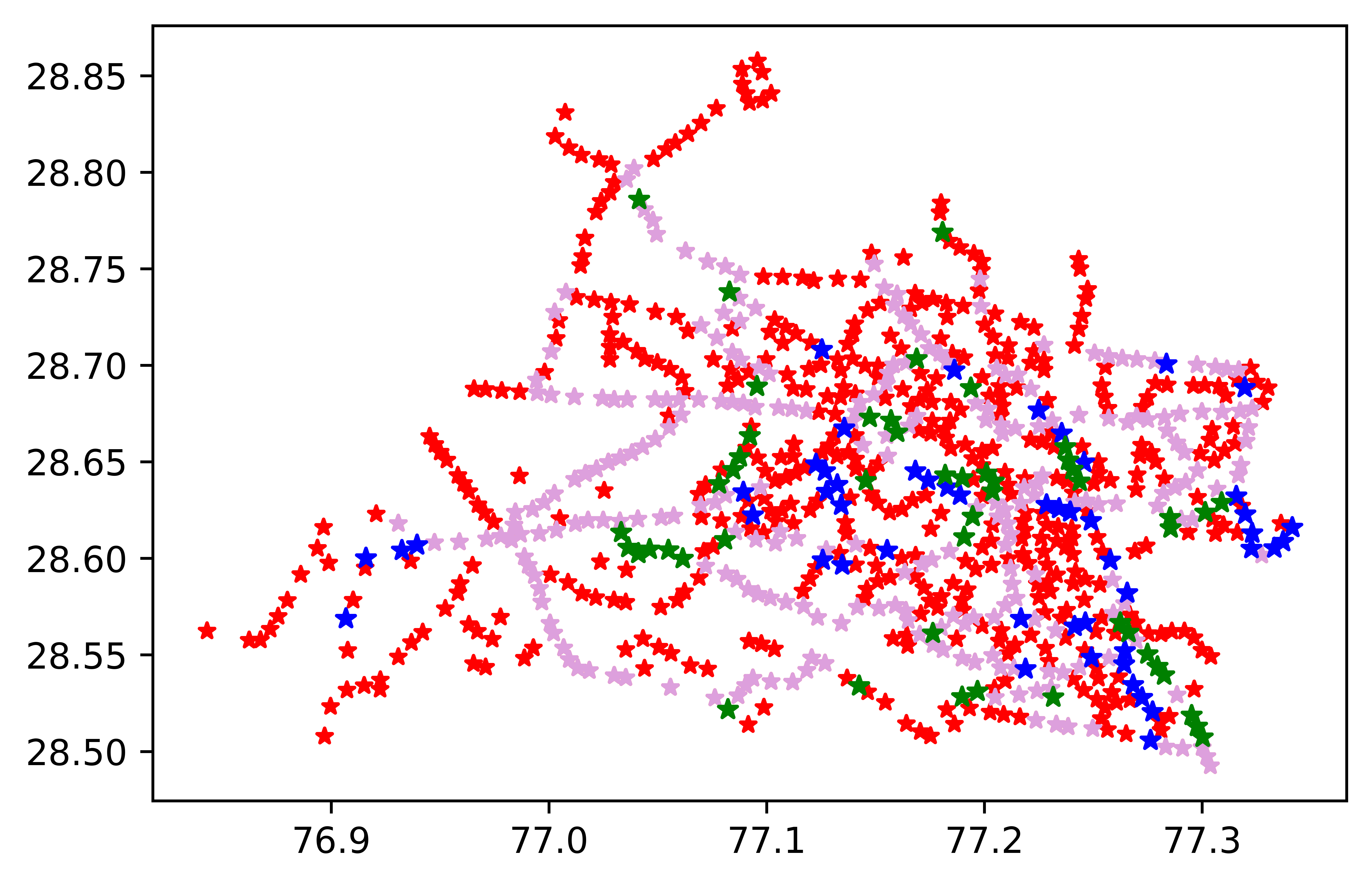}
        \caption{RFL vs FLS for atleast 10 time stamp}
    \end{subfigure}
    \caption{Coverage plots for all the locations of Delhi for selected bus $k$=30. Red points denote the point not sampled by both RFL and comparison Framework, Pink points denote the points sampled by both RFL and comparison Framework, Blue points denote the points sampled by RFL but not the comparison framework, Green points denote the points sampled by comparison framework but not the RFL}
    \label{sen}
\end{figure*}

To generate a spatiotemporal matrix $\mathbf{Y}$ that varies smoothly over space and time, we use two variants described in the following sections. 
\subsubsection{Simulated Data 1}
\begin{itemize}
	\item 	We first factorize the actual matrix $\Z \in \mathbf{R}^{L\times T}$as 
	\[\Z= \A\Bi^{T}\]
	where $\A \in \mathbf{R}^{L \times r}$ and $\Bi \in \mathbf{R}^{T \times r}$. 
	\item We construct the matrix $\A$ for the locations using the $\G$ and a bandlimited signal.
	The similarity matrix $\G$ can be decomposed as
	\[\G= \U \Sigma \U^{T}\]
	We generate the matrix $\A$ with rank $r$ as defined in Eq. \ref{eq:related1} as. 
	
	\[\a_i= \U_{(m)} \hat{\a_i} \,\,\,\, (\forall i =1  \,\,\,\text{to} \,\,\,r) \]
	where $a_i$ represents the $i^{th}$ column of matrix $\A$, $a_i$ is sampled from a random normal distribution with standard deviation (0.5) .
	\item Similarly we construct the matrix $\Bi$ for time stamps using $\H$ and a bandlimited signal. 
	\[\H=\V 	\Lambda \V^{T} \]
	\[\b_i= \V_{(n)} \hat{\b_i} \,\,\,\, (\forall i =1  \,\,\,\text{to} \,\,\,r)\]	
	\item The overall data $\mathbf{Y}$ is generated using $\A$, $\Bi$ and noise signal $n$ distributed as zero mean and std of 0.001. 
	\[\mathbf{Y}= \A\Bi^{T} +\mathbf{N}\]
	
\end{itemize}
We run the experiments for randomly generated 30 spatiotemporal matrix $\mathbf{Y}$, where $m$ is randomly chosen from 5 to 15, $n$ is chosen from 5 to 15, $r$ is chosen randomly from 20 to 30.
\subsubsection{Simulated Data 2}
To generate a smooth signal over space and time we used the framework described in paper \cite{sti1}. 

\begin{equation}
\a_t= \U_{(m)} \hat{\a_t}
\end{equation}
The observed signal can be generated by the following Eqs:
\begin{equation}
\mathbf{y}_t= \z_t +\n_t
\end{equation}
\begin{equation}
\mathbf{z}_t=\mathbf{R} \z_{t-1} +\a_t
\end{equation}
The state transition matrix $\mathbf{R}$ is defined as a general diagonal matrix $\mathbf{R}=diag(c_1,c_2, \dots c_T)$,  where each $c$ represents the autocorrelation coefficient that describes the time correlation of the data with the delayed (one time lag) data.We randomly generate 30 spatiotemporal matrix with $c$ as 1, $m$ is randomly chosen from 5 to 15 and $r$ is chosen randomly from 20 to 30.
\begin{table*}[h!]
	\centering
	\small
	\begin{tabular}{|l|l|l|l|l|l|l|l|l|l|l|l|l|}
		\hline
		& \multicolumn{4}{c|}{$k= 20$ } &  \multicolumn{4}{c|}{ $k=35$}& \multicolumn{4}{c|}{$k=50$ }   \\ \hline
		& PSC & PC & FLS & RFL & PSC & PC & FLS & RFL & PSC & PC & FLS & RFL \\ \hline
		Random bus & 56.796 & 4.612 & 90.202 & 91.936 & 59.102 & 6.593 & 90.379 & 91.563 & 64.078 & 10.584 & 92.515 & 93.579 \\ \hline
		Max Coverage & 55.218 & 7.357 & 91.867 & 93.163 & 68.932 & 11.983 & 93.801 & 94.693 & 79.49 & 16.117 & 95.089 & 95.764 \\ \hline
		Max Cov Loc & 80.704 & 5.141 & 90.719 & 92.593 & 92.233 & 7.756 & 92.734 & 94.143 & 96.359 & 10.293 & 93.935 & 94.941 \\ \hline
		FLS & 60.316 & 6.279 & 93.164 & 94.159 & 73.908 & 10.685 & 94.924 & 95.673 & 75.728 & 14.482 & 95.804 & 96.363 \\ \hline
		RFL($\rho=0.95$) & 58.981 & 6.349 & 93.122 & 94.201 & 71.845 & 10.628 & 94.885 & 95.654 & 75.85 & 14.629 & 95.785 & 96.392 \\ \hline
		RFL($\rho=0.98$)  & 60.68 & 6.504 & 93.16 & 94.324 & 72.816 & 10.761 & 94.875 & 95.724 & 79.369 & 14.759 & 95.754 & 96.414 \\ \hline
		RFL($\rho=0.99$)  & 64.442 & 6.394 & 92.842 & 94.223 & 73.908 & 10.68 & 94.812 & 95.691 & 79.126 & 14.525 & 95.665 & 96.385 \\ \hline
		RFL($\rho=1$)  & 71.845 & 4.949 & 91.019 & 92.795 & 86.529 & 8.628 & 93.375 & 94.614 & 93.204 & 12.141 & 94.472 & 95.443 \\ \hline
	\end{tabular}
	\caption{Performance Comparison for selecting $k$ buses}
	\label{tab:1}
\end{table*}

\begin{table*}[h!]
	\footnotesize
	\begin{tabular}{|l|l|l|l|l|l|l|l|l|l|l|l|l|l|l|}
		\hline
		& \multicolumn{7}{c|}{Simulated Data 1 } &  \multicolumn{7}{c|}{Simulated Data 2} \\ \hline
		$k$ & Random & MC & MCL & FLS & RFL95 & RFL98 & RFL1 & Random & MC & MCL & FLS & RFL95 & RFL98 & RFL1 \\ \hline
		20 & 70.595 & 39.333 & 55.582 & 31.71 & 31.395 & 28.507 & 55.577 & 70.239 & 40.13 & 40.225 & 27.68 & 28.03 & 25.85 & 43.70 \\ \hline
		35 & 47.062 & 22.29 & 27.938 & 10.69 & 11.065 & 10.664 & 21.209 & 40.26 & 19.906 & 16.293 & 8.395 & 9.433 & 8.563 & 13.75 \\ \hline
		50 & 31.6 & 14.193 & 16.62 & 7.465 & 7.367 & 6.983 & 13.997 & 26.674 & 11.699 & 10.822 & 5.228 & 5.62 & 4.251 & 8.126 \\ \hline
		75 & 23.546 & 11.275 & 11.296 & 5.312 & 5.07 & 5.065 & 10.832 & 19.46 & 8.448 & 7.197 & 2.736 & 2.627 & 2.532 & 7.263 \\ \hline
		100 & 18.844 & 10.375 & 8.65 & 4.289 & 4.271 & 4.112 & 7.517 & 14.732 & 8.035 & 5.785 & 1.973 & 1.817 & 1.79 & 4.487 \\ \hline
	\end{tabular}
	\caption{MRE for Dense Map using VBMC(CS) }
	\label{tab:2}
\end{table*}
\begin{table*}[h!]
	\centering
	\footnotesize
	\begin{tabular}{|l|l|l|l|l|l|l|l|l|l|l|l|l|l|l|}
		\hline
			& \multicolumn{7}{c|}{Simulated Data 1 } &  \multicolumn{7}{c|}{Simulated Data 2} \\ \hline
		$k$ & Random & MC & MCL & FLS & RFL95 & RFL98 & RFL1 & Random & MC & MCL & FLS & RFL95 & RFL98 & RFL1 \\ \hline
		20 & 90.788 & 37.538 & 73.922 & 36.585 & 36.749 & 33.602 & 76.477 & 77.619 & 35.275 & 53.419 & 30.424 & 31.479 & 28.864 & 57.799 \\ \hline
		35 & 44.446 & 21.008 & 26.995 & 10.934 & 11.609 & 10.124 & 19.386 & 39.429 & 20.8 & 19.972 & 9.817 & 11.193 & 9.297 & 14.858 \\ \hline
		50 & 27.766 & 13.303 & 15.125 & 7.018 & 7.492 & 6.574 & 11.678 & 28.65 & 13.526 & 11.627 & 5.801 & 6.398 & 4.904 & 8.027 \\ \hline
		75 & 20.34 & 10.068 & 10.15 & 4.528 & 4.586 & 4.52 & 8.842 & 21.873 & 9.704 & 7.38 & 2.581 & 2.591 & 2.581 & 5.178 \\ \hline
		100 & 16.33 & 9.52 & 7.187 & 3.809 & 3.769 & 3.684 & 6.385 & 16.922 & 9.201 & 5.243 & 1.76 & 1.638 & 1.524 & 3.547 \\ \hline
		
	\end{tabular}
	\caption{MRE for Dense Map using VBSF(CS)}
	\label{tab:3}
\end{table*}
\subsection{Creating Dense Maps from Sampled Data}

Dense AQ map can be created by first selecting the $\M$ set of buses then sampling the data $\mathbf{\hat{Y}}$ from the simulated ground truth AQ matrix $\mathbf{Y}$ based on the sampling matrix $\X (\M)$. The final step is to impute the missing data in the matrix $\mathbf{\hat{Y}}$. To evaluate the performance, we compute the MRE score for the missing data imputation. 
There exist a problem of cold start while imputation, i.e., some of the locations are not sampled at all by the selected set of buses. Therefore, we use an extended version of the matrix imputation method that handles cold start cases. The matrix completion framework (VBMC)  proposed in paper \cite{mi1} imposes a low rank structure on the data to impute the missing data as:
\begin{align}
\L_1= \min_{\A,\Bi}|| \mathbf{P}_{\Omega}  (\mathbf{Y}-\A \Bi^{T})||_{{F}} \label{m1}
\end{align}
where $\A \in \mathcal{R}^{L \times r}$ and $\Bi \in \mathcal{R}^{T \times r}$ and $r=$rank$(\mathbf{Y}) << min(L,T)$
Further, VBSF \cite{ownwork} add a regularization on the matrix $\Bi$ to incorporate the temporal evolution in addition to the low rankness (Eq. \ref{m1}) as 
\begin{align}
\mathcal{R}(\Bi)= \sum_{i=1}^{T}||\b_{i}-\mathbf{F}\b_{i-1}||  \label{m2}
\end{align}


However, these framework does not incorporate for the cold start locations. Incorporating the similarity matrix $\G$ along with the low rank matrix completion framework can tackle the cold start problem \cite{cold1, cold2}. 
\begin{align}
\L_2= \min_{\A,\C}|| (\mathbf{G}-\A \C^{T})||_{{F}} \label{m3}
\end{align}
We use Variational Bayesian Matrix Completion (Cold start) and Variational Bayesian Subspace Filtering (Cold start) as the extended matrix completion frameworks that handles the cold start location data imputation to evaluate the performance of the dense AQ maps. 
We impute the missing data using VBMC(CS)  where we optimize the Eqs. (\ref{m1}, \ref{m3}). We also impute the missing data using VBSF(CS)  where we optimize the Eqs. (\ref{m1}, \ref{m2}, \ref{m3}).

\subsection{Dataset}
We use the Delhi Bus GTFS data \cite{otd} to obtain the bus occupancy tensor defined by ${\Y} \in \{0,1\}^{L\times T\times B}$  
\begin{itemize}
	\item \textbf{Locations} ($L$): There are total 3210 bus stops. We sample $L$ stops such that the minimum distance between the stops should be $d$. The total number of locations $L$ is 824 when $d=500$ m.
	\item  \textbf{Time stamps} (T): We use 10 min sampling from 6 am to 10 pm, resulting in $T$=96 as the buses run during this time. 
	\item \textbf{Buses} ($B$): The number of buses $B$  is 1476. 
\end{itemize}
The entries in $y_{l,t,b}=1$ if bus $b$ is in the 500 meter radius of location $l$ for the time stamp $t$.\\

\subsection{Evaluation Metrics}
The evaluation metrics used for the performance comparison are defined as
\begin{itemize}
	\item Percentage Stop coverage(PSC) is defined in Eq. \eqref{eq:PSC}. 
	\item Percentage Coverage (PC) is defined in Eq. \eqref{eq:PC} 
	\item FLS gain defined in Eq. \ref{eq:FLS}
	\item RFL gain defined in Eq. \ref{eq:RFL} for $\rho=0.98$
	\item Mean Relative Error (MRE): $\frac{\norm{\mathbf{Y}-\hat{\mathbf{Y}}}}{\norm{\mathbf{Y}}} $, where $\hat{\mathbf{Y}}$ is the estimate of $\mathbf{Y}$. 
	
\end{itemize}
Mean Relative Error is used to evaluate the performance of imputed dense AQ maps. 
\subsection{Performance Comparison}

\subsubsection{Effect of $\rho$ in RFL: }
As the $\rho$ increases, PSC increases. For $\rho=1$, the performance is comparable to the MCL, since if a location is sampled for a timestamp, the gain corresponding to sampling other timestamps will be zero; therefore MCL will select the buses that sample different locations to increase the Percentage stop coverage (PSC). However, the performance of RFL ($\rho=1$) worsens for PC, FLS and RFL. For $\rho=0.95$, the performance is comparable to the FLS as the effect of temporal correlation decreases with time in RFL for lower values of $\rho$. Since the FLS does not incorporate the temporal correlation in the framework and can sample the same set of representative locations across timestamps, we observe a reduced PSC for RFL ($\rho=0.95$) and FLS. 
We observe similar performance for RFL($\rho=0.98$) and RFL($\rho=0.99$). However we observe an improved FLS score for RFL($\rho=0.98$) through out as compared to RFL($\rho=0.99$). Therefore we evaluate the performance of RFL for $\rho=0.98$ rather than $\rho =0.99$ for dense map creation. 
\subsubsection{Coverage Plot:}
To demonstrate the representations in space and time, we plot the coverage plot as shown in Fig \ref{sen}.  Every point on the plot is one of the bus stop locations. Fig  \ref{sen} (a-c) denotes the locations that are sampled for atleast one timestamp, thereby illustrating the overall stops coverage. Fig \ref{sen} (d-f) represents the locations that are sampled for atleast 10 timestamps, thereby demonstrating the temporal coverage for the sampled stops. Pink points denote the locations sampled by both the RFL and the compared framework. Blue points denote the point sampled by the RFL but not the compared framework, green points denote the points sampled by the comparison framework but not the RFL and red points denote the points not sampled by RFL and the compared framework.
From Fig. \ref{sen}(a,d), it is observed that the MC does not sample a diverse set of locations. MCL provides improved spatial coverage, however the temporal coverage is worse as compared to RFL as shown in Fig. \ref{sen}(b,e). It can also be observed that RFL samples a diverse set of locations than FLS as shown in Fig \ref{sen}(c,f).
\subsubsection{Dense AQ map MRE:}
The MRE scores for creating dense AQ map is shown in Table \ref{tab:2}, \ref{tab:3}. For simplicity we denote the RFL($\rho=0.95$) as RFL95. It is observed that RFL($\rho=0.98$) outperforms all the baseline algorithms for dense map creation VBMC(CS) and VBSF(CS). Since the generated data have temporal correlation thereby VBSF(CS) performance is better than VBMC(CS) for most of the cases, especially for higher bus sampling. However, for the lower sampling of buses (20), it is observed that VBMC(CS) performs better as there are significantly lower samples to learn the temporal pattern along with the low rankness. \par 
All the experiments are run on Matlab/Python with the system configuration of 2.3 GHz and 16 GB RAM.
\section{Conclusion}
The anytime-anywhere AQ map is the holy grail of AQ monitoring and is crucial for combating air pollution, especially in the developing world. To obtain a dense air quality map, the cities may require only a small fraction of moving sensors as compared to all static sensors setup. The spatiotemporal correlation in the air quality data can be leveraged to select the moving sensor for sampling and to facilitate an effective spatiotemporal extrapolation on the sampled data. This paper proposes a Regressive Facility Location for drive-by sensing to select the set of buses that samples the spatiotemporal AQ data. It is shown that the chosen set of buses provides representative coverage across space and time. We further obtain the dense AQ maps using the matrix completion framework from the sampled spatiotemporal data and observe that RFL provides more accurate dense AQ maps. 

%% file: supplementary.tex




\section{Appendix}
\subsection{Proof of Theorem 1}
Let $\samp_t$ be the set of locations that corresponds sampled locations in the sampling matrix $\X$ for time stamp (column) $t$. The FLS gain is defined as:
\begin{equation}
FLS (\M)= \frac{\sum\limits_{ t \in \T} \sum\limits_{l\in \L} \pi^{l}_{t} (\M )}{ L \times T} 
\end{equation}
where
\begin{equation}
\pi^{l}_{t} (\M )=\max\limits_{\forall m \in \samp_t (\M)}(\S_{l,m} )
\end{equation}

\subsubsection{Monotone non decreasing }
A function f is submodular if $\forall \,\,\,\C \subseteq \D$
\[f(\C) \le f(\D) \]
Proof:\\
Let $\C$ and $\D$ be the subset of buses following $\forall \,\,\,\C \subseteq \D \subseteq \B$. 
We define $\X(\C)$ and $\X(\D)$ as the sampling matrix for the subset of buses $\C$ and $\D$. \\
Let $\samp_t$ be the set of locations that corresponds to 1 in the sampling matrix $\X(\C)$ for the time stamp $t$ and Let $\samp_t^{'}$ be the set of locations that corresponds to 1 in the sampling matrix $\X(\D)$ for the  time stamp $t$. \\
\[ \samp_t = \{i\} \,\,\,\forall \,\,\,\X_{i,t}(\C)|_{=1}\]
\[ \samp_t^{'} = \{i\} \,\,\,\forall \,\,\,\X_{i,t}(\D)|_{=1}\]
Since $\C \subseteq \D$, therefore $\samp_t \subseteq \samp_t^{'}$  $\forall \,\,\,t =1 \,\,\,\text{to} \,\,\,T$.\par

For all $l \in \L$, $ t \in \T $ we have
\begin{equation}
\max\limits_{\forall m \in \samp_t}  (\S_{l,m}) \le \max\limits_{\forall m \in \samp_t^{'}}(\S_{l,m} )
\label{eq:mono1}
\end{equation}

Hence $	f(\C) \le f(\D) $

\subsubsection{Submodular}

A function f is submodular if $\forall \,\,\,\C \subseteq \D \subseteq \B$ and $b \in \B \backslash \D$
\[f(\C \cup b)-f(\C) \ge f(\D \cup b)-f(\D)  \]

Consider a bus $b \in \B \backslash \D$, We add the bus $b$ in both $\C$ and $\D$ and analyse the gain. 
Let the bus traverse the set of location $G_t$ for a time stamp $t$. \\
\[ G_t = \{i\} \,\,\,\forall \,\,\,\X_{i,t}(b)|_{=1}\]

For all $l \in L$ and $t =1 \,\,\,\text{to} \,\,\,T$ we have
\begin{align}
\pi_{t}^l(\C \cup b)-\pi_{t}^l(\C)=\max\limits_{\forall m \in \samp_t \cup G_t}(\S_{l,m} )  - \max\limits_{\forall m \in \samp_t}(\S_{l,m} ) &\nonumber\\= \max (0,\,\,\,\max\limits_{\forall m \in G_t}(\S_{l,m} )  - \max\limits_{\forall m \in \samp_t}(\S_{l,m} )  )
\label{eq:proof1}
\end{align}
Since \[\max\limits_{\forall m \in \samp_t}(\S_{l,m} )  \le \max\limits_{\forall m \in \samp_t^{'}}(\S_{l,m} )\]
Therefore,
\begin{align}
\pi_{t}^l(\C \cup b)-\pi_{t}^l(\C)=\max (0,\,\,\,\max\limits_{\forall m \in G_t}(\S_{l,m} )  - \max\limits_{\forall m \in \samp_t}(\S_{l,m} )  ) &\nonumber\\\ge  \max (0,\,\,\,\max\limits_{\forall m \in G_t}(\S_{l,m} )  - \max\limits_{\forall m \in \samp_t^{'}}(\S_{l,m} )  )
\label{eq:proof2}
\end{align}

\begin{equation}
\pi_{t}^l(\C \cup b)-\pi_{t}^l(\C) \geq \max (0,\,\,\,\max\limits_{\forall m \in G_t}(\S_{l,m} )  - \max\limits_{\forall m \in \samp_t^{'}}(\S_{l,m} )  )
\label{eq:proof3}
\end{equation}
\begin{equation}
\pi_{t}^l(\C \cup b)-\pi_{t}^l(\C) \geq	\max\limits_{\forall m \in \samp_t^{'} \cup G_t}(\S_{l,m} )  - \max\limits_{\forall m \in \samp_t^{'}}(\S_{l,m} ) 
\label{eq:proof4}
\end{equation}
\begin{equation}
\pi_{t}^l(\C \cup b)-\pi_{t}^l(\C) \geq \pi_{t}^l(\D \cup b)-\pi_{t}^l(\D) 
\label{eq:proof5}
\end{equation}

From Eqs (\ref{eq:proof1}-\ref{eq:proof5}) for all $l \in \L$ and $t =1 \,\,\,\text{to} \,\,\,\T$ we have, 

\[f(\C \cup b)-f(\C) \ge f(\D \cup b)-f(\D)  \]

\subsection{ Prove for Corollary 1}
FLST gain defined via Eq. \ref{eq:piflst} is equivalent to the RFL gain defined via Eq. \ref{eq:pirfl}. 

\begin{equation}
\lambda_{t}^{l}(\M)=\max\limits_{ m, j\in \X (\M)} (\S_{l,m} *\mathbf{T}_{t,j})
\label{eq:piflst}
\end{equation}
\begin{equation}
\mathbf{T}_{t,j}= 
\begin{cases}
\rho^{t-j},&  t\geq j\\
0,              & t< j
\end{cases}
\end{equation}
\begin{equation}
\pi_{t}^{l}(\M)=\max((\S_{l,m})_{\forall m \in \samp_t(\M)} , \rho \pi_{t-1}^{l}(\M))  \,\,\,\forall l \in \L
\label{eq:pirfl}
\end{equation}
Proof:\\
For all $l \in \L$ and $t =1 $ using Eq. \ref{eq:piflst} we have, 
\begin{equation}
\lambda_{1}^{l}(\M)=\max( \max(\S_{l,m})_{\forall m \in \samp_1(\M)} ,V)  \,\,\,\forall l \in \L
\end{equation}
\[V=\max (\S_{l,m})_{\forall m \in \samp_j(\M)}   \,\,\,\,(\forall j >1) =0\]

Therefore, 
\begin{equation}
\lambda_{1}^{l}(\M)=\max( (\S_{l,m})_{\forall m \in \samp_1(\M)} )  \,\,\,\forall l \in \L
\end{equation}
\[\lambda_{1}^{l}(\M)=\pi_{1}^{l}(\M)\]

For all $l \in \L$ and $t =2 $   we have, 
\begin{equation}
\lambda_{2}^{l}(\M)=\max( \max(\S_{l,m})_{\forall m \in \samp_2(\M)} ,\max ((\S_{l,m})_{\forall m \in \samp_1(\M)})*\rho , 0)  
\end{equation}

\begin{equation}
\lambda_{2}^{l}(\M)=\max( \max(\S_{l,m})_{\forall m \in \samp_2(\M)} ,\pi_{1}^{l}(\M)*\rho ) 
\end{equation}
\begin{equation}
\lambda_{2}^{l}(\M)=\max((\S_{l,m})_{\forall m \in \samp_2(\M)} , \rho \pi_{1}^{l}(\M))  = \pi_{2}^{l}(\M)  
\end{equation}

For $l \in \L$ and $t =n $  suppose it is true that  $\lambda_{n}^{l}(\M))  = \pi_{n}^{l}(\M) $ and,

\begin{align}
\lambda_{n}^{l}(\M)=\max( &\max(\S_{l,m})_{\forall m \in \samp_n(\M)} ,
\nonumber \\&\max ((\S_{l,m})_{\forall m \in \samp_{n-1}(\M)})*\rho , \dots \max ((\S_{l,m})_{\forall m \in \samp_1(\M)})*\rho^{n-1})  
\label{ref:coll1}
\end{align}

For $l \in \L$ and $t =n+1 $ we have, 
\begin{align}
\lambda_{n+1}^{l}(\M)=\max&( \max(\S_{l,m})_{\forall m \in \samp_{n+1}(\M)}, \nonumber \\&  \max ((\S_{l,m})_{\forall m \in \samp_{n}(\M)})*\rho , \dots \max ((\S_{l,m})_{\forall m \in \samp_1(\M)})*\rho^{n})  
\end{align}

From Eq. \ref{ref:coll1}, 
\begin{equation}
\lambda_{n+1}^{l}(\M)=\max( \max(\S_{l,m})_{\forall m \in \samp_{n+1}(\M)} ,\rho*\lambda_{n}^{l}(\M))  
\end{equation}
Hence, 
\begin{equation}
\lambda_{n+1}^{l}(\M)=\max((\S_{l,m})_{\forall m \in \samp_{n+1}(\M)} , \rho \pi_{n}^{l}(\M))  = \pi_{n+1}^{l}(\M)  
\end{equation}

\subsection{Prove of Theorem 2}

Let $\samp_t$ be the set of locations that corresponds to sampled locations in the sampling matrix $\X$ for time stamp $t$. For $\pi_{0}=0$, $\rho \in [0,1]$, the regressive facility location gain is defined by 
\begin{equation}
RFL(\M )=  \frac{  \sum\limits_{ t \in \T} \sum\limits_{l \in \L}\pi_{t}^{l}(\M) }{L \times T}
\label{eq:RFL}
\end{equation}
where
\begin{equation}
\pi_{t}^{l}(\M)=\max((\S_{l,m})_{\forall m \in \samp_t(\M)} , \rho \pi_{t-1}^{l}(\M))  \,\,\,\forall l \in \L
\label{eq:pirfl}
\end{equation}

\subsubsection{Monotone non decreasing }
A function f is submodular if $\forall \,\,\,\C \subseteq \D \subseteq \B$
\[f(\C) \le f(\D) \]
Proof:\\
Let $\C$ and $\D$ be the subset of buses following $\forall \,\,\,\C \subseteq \D \subseteq \B$. 
We define $\X(\C)$ and $\X(\D)$ as the sampling matrix for the subset of buses $\C$ and $\D$. \\
Let $\samp_t$ be the set of locations that corresponds to 1 in the sampling matrix $\X(\C)$ for the time stamp $t$ and Let $\samp_t^{'}$ be the set of locations that corresponds to 1 in the sampling matrix $\X(\D)$ for the  time stamp $t$. \\
\[ \samp_t = \{i\} \,\,\,\forall \,\,\,\X_{i,t}(\C)|_{=1}\]
\[ \samp_t^{'} = \{i\} \,\,\,\forall \,\,\,\X_{i,t}(\D)|_{=1}\]

For all $l \in \L$ and $t =1 $ we have
\begin{equation}
\max\limits_{\forall m \in \samp_1}(\S_{l,m} )  \le \max\limits_{\forall m \in \samp_1^{'}}(\S_{l,m} )
\label{eq:mono2}
\end{equation}

For $t=2$ we have from using Eq. (\ref{eq:mono1}), 
\begin{equation}
\max((\S_{l,m})_{\forall m \in \samp_2} , \rho \max\limits_{\forall m \in \samp_1}(\S_{l,m} ) )  \le \max((\S_{l,m})_{\forall m \in \samp_2^{'}} , \rho \max\limits_{\forall m \in \samp_1^{'}}(\S_{l,m} ) )
\end{equation}

for time stamp $t$ we have, from using Eq. (\ref{eq:mono1}) 
\begin{align}
\max((\S_{l,m})_{\forall m \in \samp_t} , \rho \max\limits_{\forall m \in \samp_{t-1}}(\S_{l,m} ), \dots ,\rho^{t-1} \max\limits_{\forall m \in \samp_{1}}(\S_{l,m} ))&\nonumber\\  \le \max((\S_{l,m})_{\forall m \in \samp_t^{'}} , \rho \max\limits_{\forall m \in \samp_{t-1}^{'}}(\S_{l,m} ), \dots ,\rho^{t-1} \max\limits_{\forall m \in \samp_{1}^{'}}(\S_{l,m} )) 
\label{eq:mono3}
\end{align}

Hence $	f(\C) \le f(\D) $

\subsubsection{Submodular}

A function f is submodular if $\forall \,\,\,\C \subseteq \D \subseteq \B$ and $b \in \B \backslash \D$
\[f(\C \cup b)-f(\C) \ge f(\D \cup b)-f(\D)  \]

Consider a bus $b \in \B \backslash \D$, We add the bus $b$ in both $\C$ and $\D$ and analyse the gain. 
Let the bus traverse the set of location $G_t$ for a time stamp $t$. \\
\[ G_t = \{i\} \,\,\,\forall \,\,\,\X_{i,t}(b)|_{=1}\]
For all $l \in \L$ and $t =1 \,\,\,\text{to} \,\,\,\T$ we have

\begin{align}
&\pi_{t}^l(\C \cup b)-\pi_{t}^l(\C)=\nonumber \\
& \max((\S_{l,m})_{\forall m \in \samp_t \cup G_t} , \rho \max\limits_{\forall m \in \samp_{t-1} \cup G_{t-1}}(\S_{l,m} ), \dots ,\rho^{t-1} \max\limits_{\forall m \in \samp_1 \cup G_1}(\S_{l,m} ))-\nonumber \\
&    \max((\S_{l,m})_{\forall m \in \samp_t} , \rho \max\limits_{\forall m \in \samp_{t-1}}(\S_{l,m} ), \dots ,\rho^{t-1} \max\limits_{\forall m \in \samp_{1}}(\S_{l,m} )) 
\label{eq:rfc1}
\end{align}
Let \[K=\max((\S_{l,m})_{\forall m \in \samp_t} , \rho \max\limits_{\forall m \in \samp_{t-1}}(\S_{l,m} ), \dots ,\rho^{t-1} \max\limits_{\forall m \in \samp_{1}}(\S_{l,m} ))\]
\begin{align}
&\pi_{t}^l(\C \cup b)-\pi_{t}^l(\C)=\nonumber \\
&\max((\S_{l,m})_{\forall m \in \samp_t \cup G_t} , \rho \max\limits_{\forall m \in \samp_{t-1} \cup G_{t-1}}(\S_{l,m} ), \dots ,\rho^{t-1} \max\limits_{\forall m \in \samp_1 \cup G_1}(\S_{l,m} ))-K
\label{eq:rfc2}
\end{align}

\begin{align}
&\pi_{t}^l(\C \cup b)-\pi_{t}^l(\C)=\nonumber \\
&\max(0, (\S_{l,m})_{\forall m \in  G_t}-K , \rho\max\limits_{\forall m \in G_{t-1}}(\S_{l,m} ) -K, \dots ,\rho^{t-1} \max\limits_{\forall m \in  G_1}(\S_{l,m} )-K)
\label{eq:rfc3}
\end{align}
let \[K'=\max((\S_{l,m})_{\forall m \in \samp_t^{'}} , \rho \max\limits_{\forall m \in \samp_{t-1}^{'}}(\S_{l,m} ), \dots ,\rho^{t-1} \max\limits_{\forall m \in \samp_{1}^{'}}(\S_{l,m} )) \]
From Eq. \ref{eq:mono3} since $K \leq K'$
Therefore,
\begin{align}
&\pi_{t}^l(\C \cup b)-\pi_{t}^l(\C) \geq \nonumber \\
&  \max(0, (\S_{l,m})_{\forall m \in  G_t}-K' , \rho\max\limits_{\forall m \in G_{t-1}}(\S_{l,m} ) -K', \dots ,\rho^{t-1} \max\limits_{\forall m \in  G_1}(\S_{l,m} )-K')
\label{eq:rfc4}
\end{align}
\begin{align}
&\pi_{t}^l(\C \cup b)-\pi_{t}^l(\C) \geq \nonumber \\
& \max((\S_{l,m})_{\forall m \in \samp_t \cup G_t} , \rho \max\limits_{\forall m \in \samp_{t-1} \cup G_{t-1}}(\S_{l,m} ), \dots ,\rho^{t-1} \max\limits_{\forall m \in \samp_1 \cup G_1}(\S_{l,m} ))-K'
\label{eq:rfc5}
\end{align}

\begin{align}
&\pi_{t}^l(\C \cup b)-\pi_{t}^l(\C) \geq \nonumber \\
&\max((\S_{l,m})_{\forall m \in \samp_t' \cup G_t} , \rho \max\limits_{\forall m \in \samp_{t-1}' \cup G_{t-1}}(\S_{l,m} ), \dots ,\rho^{t-1} \max\limits_{\forall m \in \samp_1' \cup G_1}(\S_{l,m} ))\nonumber \\
&  - \max((\S_{l,m})_{\forall m \in \samp_t'} , \rho \max\limits_{\forall m \in \samp_{t-1}'}(\S_{l,m} ), \dots ,\rho^{t-1} \max\limits_{\forall m \in \samp_{1}'}(\S_{l,m} )) 
\label{eq:rfc6}
\end{align}
\begin{align}
\pi_{t}^l(\C \cup b)-\pi_{t}^l(\C) \geq \pi_{t}^l(\D \cup b)-\pi_{t}^l(\D)
\label{eq:rfc7}
\end{align}
From Eqs (\ref{eq:rfc1}-\ref{eq:rfc7}) for all $l \in \L$ and $t =1 \,\,\,\text{to} \,\,\,\T$ we have, 

\[f(\C \cup b)-f(\C) \ge f(\D \cup b)-f(\D)  \]

\subsection{Creating Dense Maps from Sampled Data}


We use Variational Bayesian Matrix Completion (Cold start) and Variational Bayesian Subspace Filtering (Cold start) as the extended matrix completion frameworks that handles the cold start location data imputation to evaluate the performance of the dense AQ maps. 
We impute the missing data using VBMC(CS)  where we optimize the Eqs. (\ref{m1}, \ref{m3}). We also impute the missing data using VBSF(CS)  where we optimize the Eqs. (\ref{m1}, \ref{m2}, \ref{m3}). 
The matrix completion framework (VBMC)  proposed in paper \cite{mi1} imposes a low rank structure on the data to impute the missing data as:
\begin{align}
\L_1= \min_{\A,\Bi}|| \mathbf{P}_{\Omega}  (\mathbf{Y}-\A \Bi^{T})||_{{F}} \label{m1}
\end{align}
where $\A \in \mathcal{R}^{L \times r}$ and $\Bi \in \mathcal{R}^{T \times r}$ and $r=$rank$(\mathbf{Y}) << min(L,T)$
Further, VBSF \cite{ownwork} add a regularization on the matrix $\Bi$ to incorporate the temporal evolution in addition to the low rankness (Eq. \ref{m1}) as 
\begin{align}
\mathcal{R}(\Bi)= \sum_{i=1}^{T}||\b_{i}-\mathbf{F}\b_{i-1}||  \label{m2}
\end{align}
However, these framework does not incorporate for the cold start locations. Incorporating the similarity matrix $\G$ along with the low rank matrix completion framework can tackle the cold start problem \cite{cold1, cold2}. 
\begin{align}
\L_2= \min_{\A,\Ci}|| (\mathbf{G}-\A \Ci^{T})||_{{F}} \label{m3}
\end{align}

Refer the update equations in the papers \cite{ownwork, mi1}. We showed the changed updated equations below:  

The update for $i^{th}$ column of $\A$ for the cold start matrix completion is as follows:
\begin{align}
\Xib^\A_i &=  \left(\hat{\gamma}_i\I_{r} + \hat{\beta}\sum_{\tau \in\Omega'_i}(\mub_\tau^\Bi(\mub_\tau^\Bi)^T+\Xib_{\tau,\tau}^\Bi) +\hat{\beta_1}(\mub^\Ci(\mub^\Ci)^T+\Xib^\Ci)\right)^{-1}
\end{align}
\begin{align}
	\mub^\A_i &= \Xib^\A_i(\hat{\beta}\sum_{\tau \in\Omega'_i} \mub^\Bi_\tau y_{i\tau} + \hat{\beta_1}\mub^\Ci g_{i})
\end{align}

The update for $\beta_1$ is as follows:
\begin{equation}
\hat{\beta_1} =\frac{p\,L^2}{\parallel  \mathbf{G}-\A \Ci^{T}\parallel^2_{F}}
\end{equation}
The updates Eqs. for $\Ci$ is as follows: 
\begin{align}
\Gamma &= diag(\gamma), & \Xib^{\bf C}\tt &= (<\beta_1><\bf A^TA\tt>+\Gamma)^{-1}
\end{align}
\begin{align}
<\Ci > &= <\beta_1> \G\it< \A> \Xib^{\bf C}, 
\end{align}